%% file: main.tex
\documentclass[conference]{IEEEtran}
\IEEEoverridecommandlockouts
\usepackage{cite}
\usepackage{amsmath,amssymb,amsfonts}
\usepackage{algorithmic}
\usepackage{graphicx}
\usepackage{textcomp}
\usepackage{comment}
\usepackage[caption=false,font=footnotesize]{subfig}
\usepackage{xcolor}
\def\BibTeX{{\rm B\kern-.05em{\sc i\kern-.025em b}\kern-.08em
    T\kern-.1667em\lower.7ex\hbox{E}\kern-.125emX}}
\usepackage{multirow}
\usepackage{booktabs}
\usepackage{hyperref}

\newcommand{\acronym}{E-ADLLM} 
\newcommand{\AdlLlmBase}{ADL-LLM~\cite{civitarese2025large}}
\newcommand{\AlSH}{AL~\cite{krishnan2014activity}}

\begin{document}

\title{Improving Zero-shot ADL Recognition with Large Language Models through Event-based Context and Confidence}

\author{
\IEEEauthorblockN{Michele Fiori, Gabriele Civitarese, Marco Colussi, Claudio Bettini}
\IEEEauthorblockA{\textit{Dept. of Computer Science}\\ University of Milan, Milan, Italy\\ michele.fiori@unimi.it, gabriele.civitarese@unimi.it, marco.colussi@unimi.it, claudio.bettini@unimi.it
\\
\\
\textcolor{red}{\textit{This paper is currently under review, for any inquiries please contact michele.fiori@unimi.it}}
} 
}


\maketitle

\begin{abstract}

Unobtrusive sensor-based recognition of Activities of Daily Living (ADLs) in smart homes by processing data collected from  IoT sensing devices supports applications such as healthcare, safety, and energy management. Recent zero-shot methods based on Large Language Models (LLMs) have the advantage of removing the reliance on labeled ADL sensor data. However, existing approaches rely on time-based segmentation, which is poorly aligned with the contextual reasoning capabilities of LLMs. Moreover, existing approaches lack methods for estimating prediction confidence. This paper proposes to improve zero-shot ADL recognition with event-based segmentation and a novel method for estimating prediction confidence. Our experimental evaluation shows that event-based segmentation consistently outperforms time-based LLM approaches on complex, realistic datasets and surpasses supervised data-driven methods, even with relatively small LLMs (e.g., Gemma 3 27B). The proposed confidence measure effectively distinguishes correct from incorrect predictions.


\end{abstract}


\section{Introduction and Related Work}\input{sections/introduction}


\section{Problem Formulation and Architecture}
\input{sections/formalization_architecture}

\section{Improving Through Event-based Context}
\input{sections/event_based}

\section{Improving Through Confidence}
\input{sections/confidence}

\section{Experimental Evaluation}
\input{sections/experiments}

\section{Discussion}

\input{sections/discussion}

\section{Conclusions}
\input{sections/conclusion}


\bibliographystyle{IEEEtran}
\bibliography{references}

\input{sections/appendix}

\end{document}

%% file: sections/introduction.tex

The recognition of Activities of Daily Living (ADLs) is a core capability of smart homes equipped with IoT sensing devices, enabling applications in healthcare, safety, and wellbeing~\cite{chakraborty2023smart,babangida2022internet}. ADLs represent self-care activities such as cooking, eating, or taking medication, and their accurate recognition is crucial, for instance, to monitor independence and detect early cognitive decline in elderly subjects~\cite{riboni2016smartfaber}. The majority of the existing approaches for ADL recognition are based on data-driven methods (e.g., supervised, semi-supervised, self-supervised learning)~\cite {chen2012sensor,babangida2022internet,liciotti2020sequential,fahad2021activity,arrotta2023micar,chen2024utilizing}. However, such approaches require labeled training data, whose collection is costly and often prohibitive in real-world settings. This limitation affects not only fully supervised training, which relies on large labeled datasets, but also the fine-tuning of pre-trained models for a target home environment, since acquiring even a small amount of labeled data can involve significant practical and privacy challenges. As an additional limitation, data-driven models are often used as \textit{``black boxes''} and the rationale behind their activity predictions is opaque.

Considering recent related work, Large Language Models (LLMs) have been explored for zero-shot ADL recognition in smart homes~\cite{civitarese2025large,cleland2024leveraging,chen2024towards,fritsch2024hierarchical,thukral2025layout}. By transforming the raw time series obtained from IoT sensing devices into textual or structured representations and exploiting prompting strategies, these approaches remove the need for labeled data~\cite{civitarese2025large,cleland2024leveraging,gao2024unsupervised,fiori2025leveraging,xia2023unsupervised,fritsch2024hierarchical,chen2024towards}. 

Despite this flexibility, all existing LLM-based methods still rely on time-based segmentation, a design choice inherited from earlier data-driven models. 
%
In fixed time-based segmentation, activities are classified over temporal windows. While effective for sensors with constant sampling rates, selecting an appropriate window size for ambient sensor data is more challenging. Indeed, IoT sensors like PIRs (to detect motion) and magnetic (to detect opening/closing of doors and drawers) generate sparse events based on the interaction of the inhabitant with the environment. In this scenario, small temporal windows may lack sufficient context information, while large windows may mix multiple activities and introduce noise.
As a more flexible alternative, earlier work proposed event-based segmentation~\cite{krishnan2014activity}, in which each window comprises a fixed number of sensor events, regardless of their temporal spacing, and the objective is to classify the ADL associated with the last event in the window.
This approach better reflects the irregular sampling nature of smart-home sensors. However, with the advent of deep learning, most activity recognition systems reverted to time-based segmentation, as deep learning architectures typically require fixed-size time series inputs~\cite{liciotti2020sequential}.

 
An additional limitation of LLM methods is that their effectiveness has been demonstrated primarily on simple, controlled datasets, often by excluding challenging activity classes, such as \emph{other}, thereby excessively simplifying the recognition problem. As a result, these approaches show limited generalization in realistic datasets such as CASAS~\cite{cook2013casas}, where sparse, noisy sensor streams and semantically weak activity labels pose significant challenges~\cite{fiori2025leveraging}.


In this work, we argue that event-based segmentation is better aligned with the sequential reasoning capabilities of LLMs, which naturally operate on contextual histories rather than fixed temporal windows. We therefore propose \acronym{}, a novel zero-shot LLM-based ADL recognition system that infers the activity associated with each sensor event given the previous \(K\) events as context. To the best of our knowledge, this is the first study to investigate event-based context for ADL recognition using LLMs.

Another limitation of existing LLM-based approaches is the absence of reliable confidence estimates. Prompting LLMs to directly output confidence values is unreliable, but recent work shows that confidence can be inferred from response consistency across repeated queries~\cite{zhao2025learning}. Building on this insight, \acronym{} also introduces a confidence estimation method based on repeated inference, and we are the first to apply it to ADL recognition with LLMs.

Our experimental results on in-the-wild public datasets show that event-based segmentation significantly outperforms time-based LLM approaches and supervised baselines in complex scenarios, while the proposed confidence measure correlates strongly with prediction correctness.

To summarize, the contributions of our paper are the following:
\begin{itemize}
    \item We propose \acronym{}, an LLM-based ADL recognition system adopting event-based segmentation.
    \item We introduce a method to estimate confidence in LLM-based ADL predictions.
    \item Experiments on two public datasets and three language models (including open-weight ones) show that \acronym{} significantly outperforms state-of-the-art approaches.
    \item We demonstrate that the proposed confidence estimation effectively discriminates between correct and incorrect predictions.
\end{itemize}

%% file: sections/formalization_architecture.tex
\subsection{Problem Formulation}
\label{subsec:problem_formulation}
We begin by formally defining the HAR problem that this study aims to tackle.
Consistent with existing literature, we consider a smart home scenario with either a single resident or a sensing system capable of accurately identifying the person responsible for each sensor activation.

The smart home is equipped with IoT sensing devices monitoring the interaction of the inhabitant with the surrounding environment.
Each sensor generates a data stream according to its modality. For instance, binary sensors produce discrete activation/deactivation events (e.g., ON/OFF) that directly encode state changes (e.g., the fridge door is opened). Other types of sensors instead provide continuous or inertial measurements. Such streams are preprocessed to extract higher-level discrete features (e.g., high humidity, sitting posture), each represented by a start and end timestamp. The resulting multivariate event-based time series is then segmented into temporal windows for further analysis.

Let \( S = \{s_1, s_2, \ldots, s_M\} \) be the set of \(M\) sensors.
Each sensor \( s_i \) generates a stream of events, where each event is represented as a tuple:
\[e = \langle t, s_i, \sigma \rangle\]
where:
\begin{itemize}
    \item \( t \in \mathbb{R} \): the timestamp at which the event occurred
    \item \( s_i \in S \): the sensor responsible for the event
    \item \( \sigma \in \{\text{ON}, \text{OFF}\} \): the event status indicating the activation (ON) or deactivation (OFF) of the sensor or inferred high-level feature.
\end{itemize}

For example, the event
$$ e = \langle \text{2025-10-15 06:30:00}, \text{M1}, \text{ON} \rangle $$
indicates that the motion sensor M1 changed its status to ON at time 2025-10-15 06:30:00. Formally, let $ E = \{e_1, e_2, \ldots, e_T\} $ be the global event stream from all sensors, sorted by timestamp $t$. 
%

For each event $e_t \in E$, let $\mathcal{C}_t$
denote its context, defined as a subsequence of events in $E$ that may be relevant to infer the activity occurring at $e_t$. 
For example, the context may be defined by the preceding events limited by a time constraint, while in~\cite{krishnan2014activity} the context is a fixed number of previous events. The context may also exclude some events in these sequences based on the semantics of $e_t$. If real-time is not a requirement, some future events may also be considered as context.

If $A = \{a_1, a_2, \ldots, a_m\}$ is the finite set of $m$ possible ADLs, and $\mathcal{S}(E)$ is the set of all subsequences of $E$, the Human Activity Recognition (HAR) problem consists of finding a function
\[
f : E \times \mathcal{S}(E) \rightarrow A \times [0,1]
\]
such that for each event $e_t \in E$ and its context $\mathcal{C}_t \in \mathcal{S}(E)$ \, the function outputs a predicted activity and a confidence score:
\[
f(e_t, \mathcal{C}_t) = (a_t, cs_t)
\]
where
\begin{itemize}
    \item $a_t \in A$ is the activity associated with event $e_t$,
    \item $cs_t \in [0,1]$ is the confidence score associated with this prediction.
\end{itemize}

The function \(f\) can either be defined or learned.

\subsection{Overall Architecture}
\begin{figure}[]
    \centering
    \includegraphics*[width=\linewidth]{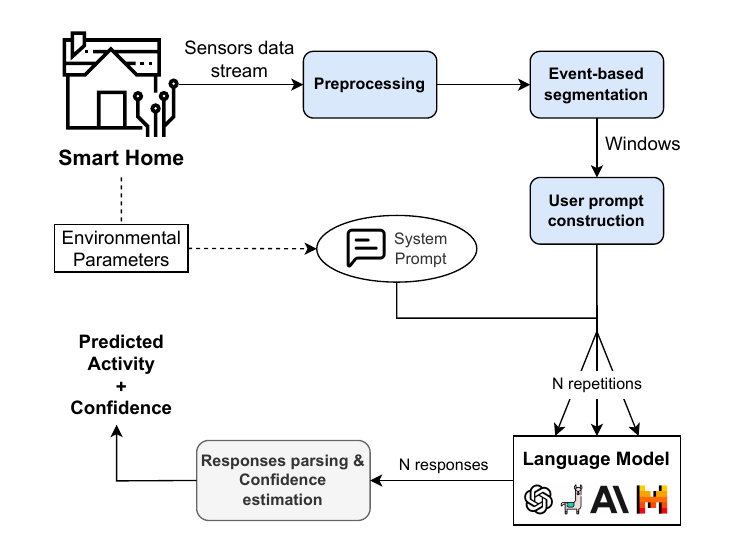}
    \caption{\acronym{} overall architecture}
    \label{fig:overall_architecture}
\end{figure}
Figure \ref{fig:overall_architecture} shows the proposed system architecture for \acronym{}: our sensor-based zero-shot ADL recognition system. 
A preprocessing module is in charge of converting raw sensor data into high-level discretized ON/OFF events. 
Moreover, it applies data cleaning to correct issues such as missing values or redundant events (e.g., consecutive ONs or consecutive OFFs). 

Once pre-processed, the event stream is segmented into 
event-based windows, where each window consists of a fixed number of events. 
Each window is converted into a textual representation (CSV) through a user prompt construction module, forming the input for the LLM. In parallel, a system prompt is used to provide additional context and constraints to guide the LLM's response.  The user and system prompts are jointly fed into an LLM, which interprets the windowed sensor data and generates a predicted activity label. Note that, consistently with the problem formulation described in~\ref{subsec:problem_formulation}, the goal of the LLM is to infer the activity that occurred during the last event, given the previous ones as context.

By prompting the LLM with the same query multiple times (i.e., $N$ repetitions), \acronym{} obtains multiple responses for the same output. All the outputs are then post-processed to infer the overall predicted ADL and the prediction confidence.

%% file: sections/event_based.tex

\subsection{Event based context}
\label{subsec:eventbased}

In this paper, we consider a simple event-based context. 
Building on top of the formalization provided in \ref{subsec:problem_formulation}, given the global event stream $E$ and an event $e_t\in E$, 
$\mathcal{C}_{t}$ is the sequence of the preceding $k-1$ events in $E$.
Then, the HAR problem becomes the problem of finding the function that, given a window $W_t$ of $k$ consecutive events
$\mathcal{C}_t \cup \{e_t\} = \{e_{t-k+1}, \ldots, e_t\}$, returns the class of activity $a_t$ predicted as occurring at $e_t$, and the confidence score $cs_t$.

Since predicting the activity being performed at each event in the stream may be too expensive and possibly useless, we apply a form of \emph{sampling} by defining a step parameter $s \in \mathbb{N}^{+}$, which controls how many events we skip for activity prediction, and consequently how many windows we consider. This is similar to deciding an overlapping factor for windows. 
%
%
More precisely, the step parameter determines the distance between consecutive windows. For $s = 1$, windows overlap by $k-1$ events, resulting in a maximally overlapping segmentation with all events being considered for prediction, whereas larger values of $s$ reduce the overlap. When $s = k$, windows are consecutive, i.e., the window following $W_{e_t}$ is $W_{e_{t+k}}$ and starts with $e_{t+1}$, while for $s > k$, some events would not be included in any window.

Note that we assume that the activity predicted for the most recently evaluated event persists until a new prediction is produced. Consequently, any skipped events are implicitly assigned the activity label predicted for that last evaluated event.

%



\subsection{System prompt}
\begin{figure}[]
    \centering
    \includegraphics*[width=\linewidth]{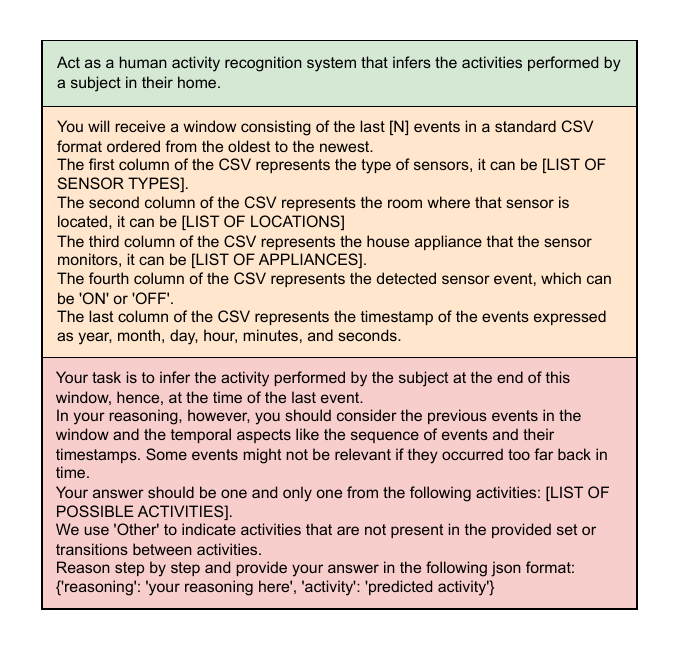}
    \caption{\acronym{}'s system prompt}
    \label{fig:system_prompt}
\end{figure}

The system prompt provides the general instructions guiding the LLM’s behavior, independently of the specific window being processed. Figure \ref{fig:system_prompt} shows our system prompt. The first section (green) sets the general behavior of the model. The second part (orange) provides a detailed description of the input format that the LLM will receive.  In addition to describing the structure of the input, this section explicitly enumerates the possible values for each attribute, sensor types, locations, and appliances, thus allowing the model to understand the layout and instrumentation of the environment. 
The third section (red) outlines the specific task the LLM must perform, that is, inferring the subject’s activity at the end of the given window, based on the sequence and timing of the events. The instructions emphasize temporal reasoning and encourage the model to ignore irrelevant or outdated events. The model is asked to select one activity from a predefined list. To guide the model’s reasoning process, it is explicitly prompted to ``reason step by step'', a formulation that aligns with the Chain-of-Thought (CoT) prompting strategy. This technique has been shown to improve LLM performance on complex tasks by encouraging intermediate reasoning steps that lead to more accurate and interpretable results \cite{wei2022chain}. While recent reasoning models may produce better CoT, we excluded them since they required too much time to generate the output.
 
Finally, the prompt enforces a JSON format for the output, which ensures that the result can be easily and reliably parsed in an automated pipeline.

\subsection{User prompt}
\begin{figure}[h]
    \centering
    \includegraphics*[width=0.7\columnwidth]{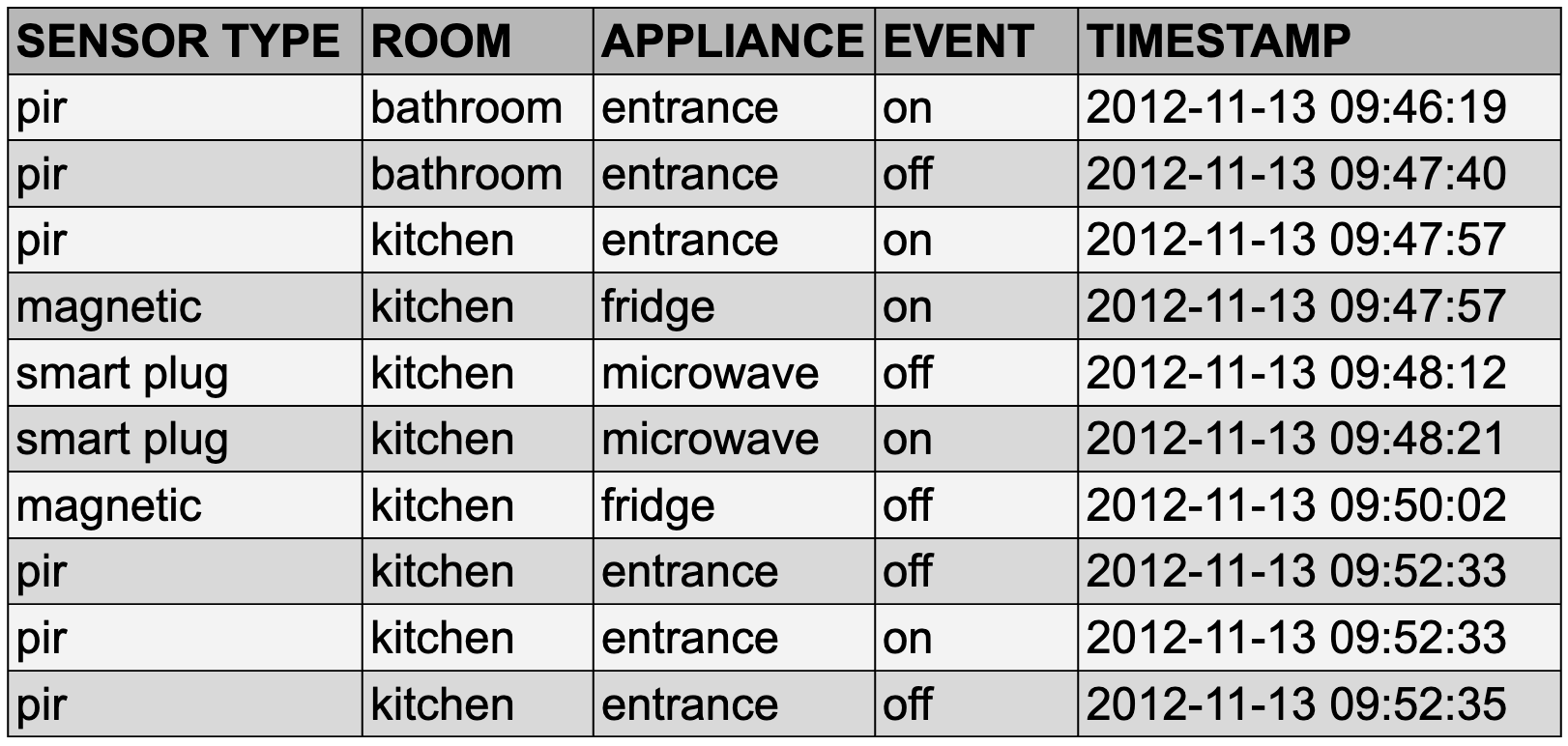}
    \caption{An example of a user prompt, represented in a tabular form for better readability.}
    \label{fig:user_prompt}
\end{figure}

Figure \ref{fig:user_prompt} shows an example of a user prompt representing the events in a window. While the data is actually represented as a CSV-formatted string, it is shown here as a table for better readability.

\vspace{3mm}

\subsection{LLM's output}
\label{subsec:llm_output}
\begin{figure}[h!]
    \centering
    \includegraphics*[width=0.8\columnwidth]{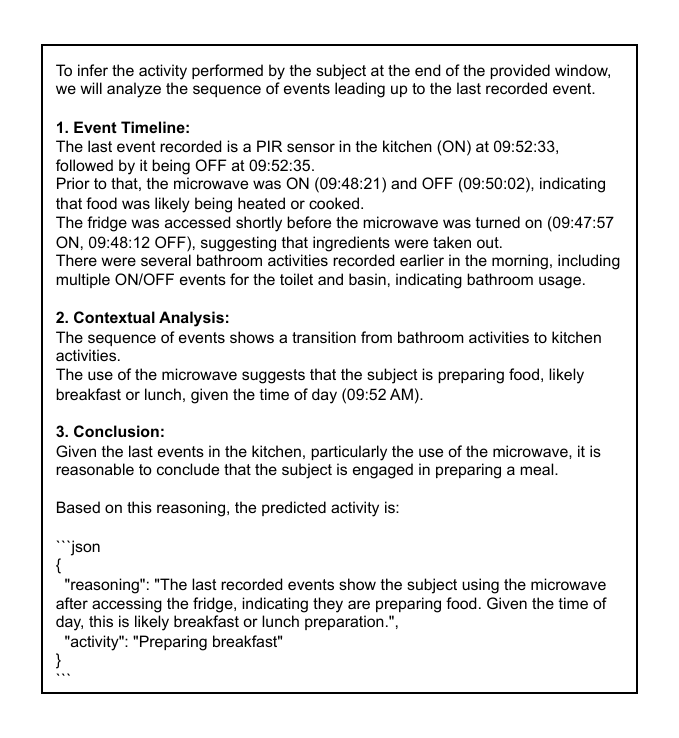}
    \caption{Output example by considering the user prompt in Figure~\ref{fig:user_prompt} as input.}
    \label{fig:output example}
\end{figure}

Figure~\ref{fig:output example}shows an example of a model’s output, including the reasoning and the final output in the standard JSON format, when the input is the one shown in Figure \ref{fig:user_prompt}.

%% file: sections/confidence.tex
\subsection{Approximate confidence for LLM's predictions}
\label{subsec:confidence}
In traditional multiclass activity recognition, data-driven models typically output a probability distribution over the set of possible activities. The predicted activity is usually the one associated with the highest probability, and this probability itself is interpreted as a measure of the model’s confidence. 
Although LLMs do internally produce token-level probabilities, using them to obtain a reliable confidence estimate at the activity level is cumbersome and impractical for activity recognition, where confidence is typically defined directly over the set of classes.
As a result, directly assessing the confidence of an LLM’s prediction is not straightforward. 
To address this issue, our methodology introduces an approximation strategy that enables the estimation of confidence scores for LLM-generated classifications. The proposed approach involves querying the LLM multiple times with the same input and aggregating the outputs. The final prediction is determined through majority voting: the activity that appears most frequently among the generated responses is selected as the predicted label. The corresponding confidence is then computed as the proportion of responses that support the majority class, i.e., the number of times the most frequently predicted activity appears divided by the total number of queries. When the number of query repetitions is lower than the number of ADL classes, there is a risk that no single class will achieve a majority. In such cases, the final prediction is resolved using an arbitrary tie-breaking strategy. This strategy can be random or based on application-driven priorities, for example, by favoring activities that are considered more important in a specific scenario. Regardless of the chosen strategy, the resulting confidence will be inherently low, equal to or close to the reciprocal of the total number of queries performed.

Formally, let:
\begin{itemize}
    \item $A = \{a_1, a_2, \dots, a_m\}$ be the finite set of $m$ possible ADLs
    \item $N$ be the number of times the LLM is prompted with the same input.
    \item $P = \{p_1, p_2, \dots, p_N\}$ be the set of $N$ activity predictions returned by the LLM, with $p_n \in A$.
    \item $freq(a)$ be the frequency count of activity $a \in A$ in $P$.
\end{itemize}

Then:
\begin{itemize}
    \item The final predicted activity $\hat{a}$ is:
    $$
    \hat{a} = \arg\max_{a \in A} freq(a)
    $$
    If multiple activities share the maximum frequency (i.e., a tie), one is selected according to the chosen strategy as described above.

    \item The approximate confidence $c(p)$ is defined as:
    $$
    c(p) = \frac{freq(\hat{a})}{N}
    $$
\end{itemize}

In the case where no activity receives a majority, $c(p)$ reflects the resulting uncertainty, with a minimum value of $\frac{1}{N}$ when all predictions are different.

%% file: sections/experiments.tex
\subsection{Datasets}

\subsubsection{CASAS Aruba}
The Aruba dataset is part of the CASAS smart home project \cite{cook2013casas}, developed by Washington State University (WSU). It contains sensor data collected from a smart home occupied by a single individual. The environment is equipped with 40 sensors distributed throughout the home, including PIR motion sensors and magnetic sensors. The dataset spans 219 days, capturing long-term daily living patterns. It covers 12 different activities: sleeping, bathroom activity, preparing meals, relaxing, housekeeping, eating, washing dishes, leaving and entering the home, working, meditating, and other.

\subsubsection{CASAS Milan}
The CASAS Milan dataset \cite{cook2013casas} is part of the CASAS project as well and has been widely adopted in the literature on activity recognition in smart homes. It contains annotated sensor data collected from the home of a single resident. The environment was instrumented primarily with motion sensors placed in different rooms to monitor resident presence, along with two temperature sensors and a small number of magnetic sensors installed on doors and drawers. The dataset spans 83 days, capturing long-term daily living patterns. It covers 10 different activities: kitchen activity, eating, bathroom activity, relaxing on armchair, sleeping, dressing/undressing, housekeeping, working, leaving home, and other.

\subsection{Dataset split}
Since this approach is inherently time-dependent, we also performed the dataset split to ensure that the test data consisted of contiguous days. Specifically, we extracted three weeks (21 days) from each dataset to serve as the test set, verifying that these selected periods provided a representative distribution of all activities. For zero-shot experiments, only the test set was used, whereas in comparisons with the supervised approach, the remaining portion of the dataset was employed as the training set.

\subsection{Metrics}
We adopt a well-known temporal-based evaluation metric for event-based segmentation~\cite{alemdar2017multi, wang2011recognizing}. Intuitively, this metric represents the percentage of the total observation time during which the predicted activity label matches the ground truth label. Note that although we may have predictions only at the timestamps of the considered events, as explained in Section \ref{subsec:eventbased}, by assuming persistence, at each timestamp we have an implicit prediction.
Technically, we consider contiguous time intervals of fixed duration $\Delta$ ($\Delta = 1s$ in the literature and in our experiments) and compare the value in the ground truth and the prediction for each interval. The accuracy of the recognition rate is the percentage of correctly predicted intervals over their total number. In order to obtain a more meaningful score, we compute and report the weighted F1 score. 


An alternative common evaluation metric in event-based HAR is instance-based evaluation, in which the predicted label of each event is compared against its ground truth label. However, instance-based metrics are time-agnostic: events are counted equally regardless of how long the state they reveal persists in time. Hence, a wrong prediction for a short interval counts the same as for a long interval.
Adopting an instance-based metric would also make the comparison against baselines based on time-based segmentation (e.g.,~\AdlLlmBase{}) problematic. Indeed, time-based segmentation implicitly weights predictions by duration.




\subsection{LLMs choice}
We considered three different LLMs with different performances, accessibility, and deployment constraints. First, we used GPT-4o mini, a state-of-the-art proprietary model developed by OpenAI, selected for its excellent trade-off between quality and costs. 
In parallel, we tested two open-source models from Google's Gemma3 family~\footnote{\url{https://deepmind.google/models/gemma/gemma-3/}}. The Gemma3 4B model
was chosen for its suitability for deployment on local edge devices or gateways, making it ideal for resource-constrained environments. Complementing this, we included the more powerful Gemma3 27B model, designed for server-grade applications such as those that may be found in a telemedicine infrastructure, where higher computational resources are available. 

\subsection{Experimental setup}
In order to experimentally evaluate our proposed system, we implemented a Python prototype. We accessed the remote OpenAI model through the official OpenAI API \footnote{https://openai.com/api/}. Regarding the open source models, instead, we deployed them locally thanks to the Ollama platform \footnote{https://ollama.com/} and their official Python package \footnote{https://pypi.org/project/ollama-python/}. The experiments were conducted on a Linux-based machine running Ubuntu 22.04.4 LTS on an x86\_64 architecture. The system is equipped with 230 GB of system memory and an NVIDIA A100 PCIe GPU with 80 GB of dedicated VRAM.

\begin{figure*}[t]
    \centering
    \subfloat[CASAS Aruba\label{fig:aruba_durations}]{
        \includegraphics[width=0.48\linewidth]{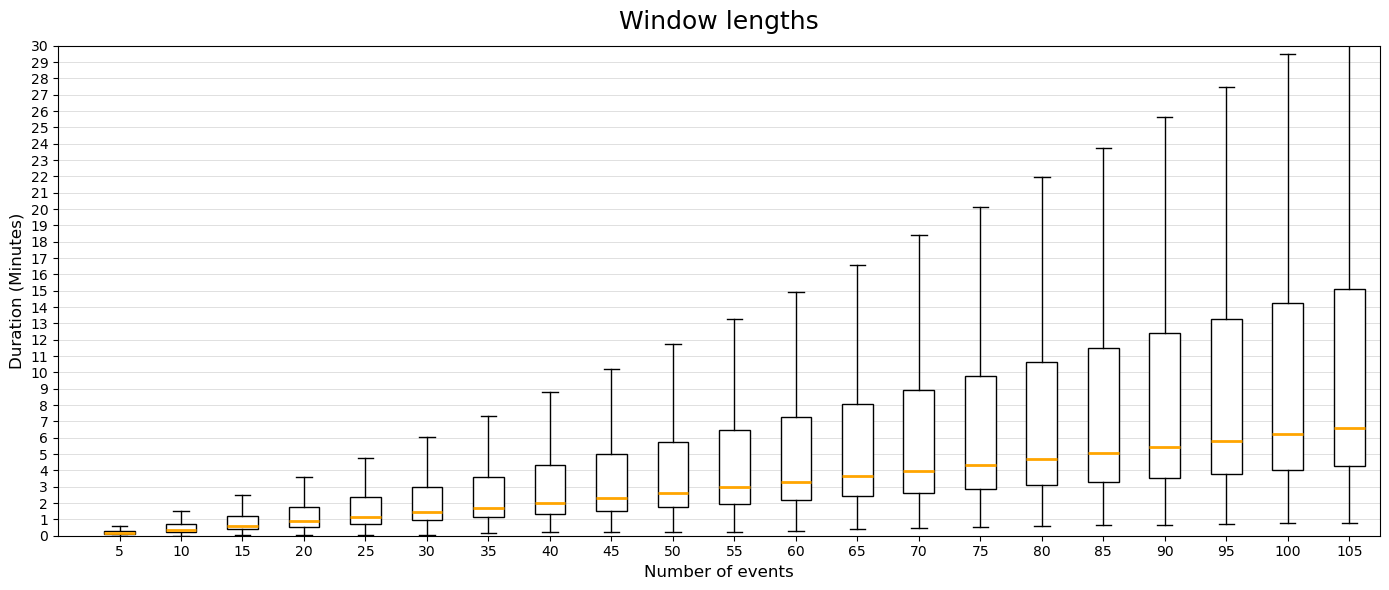}
    }\hfill
    \subfloat[CASAS Milan\label{fig:milan_durations}]{
        \includegraphics[width=0.48\linewidth]{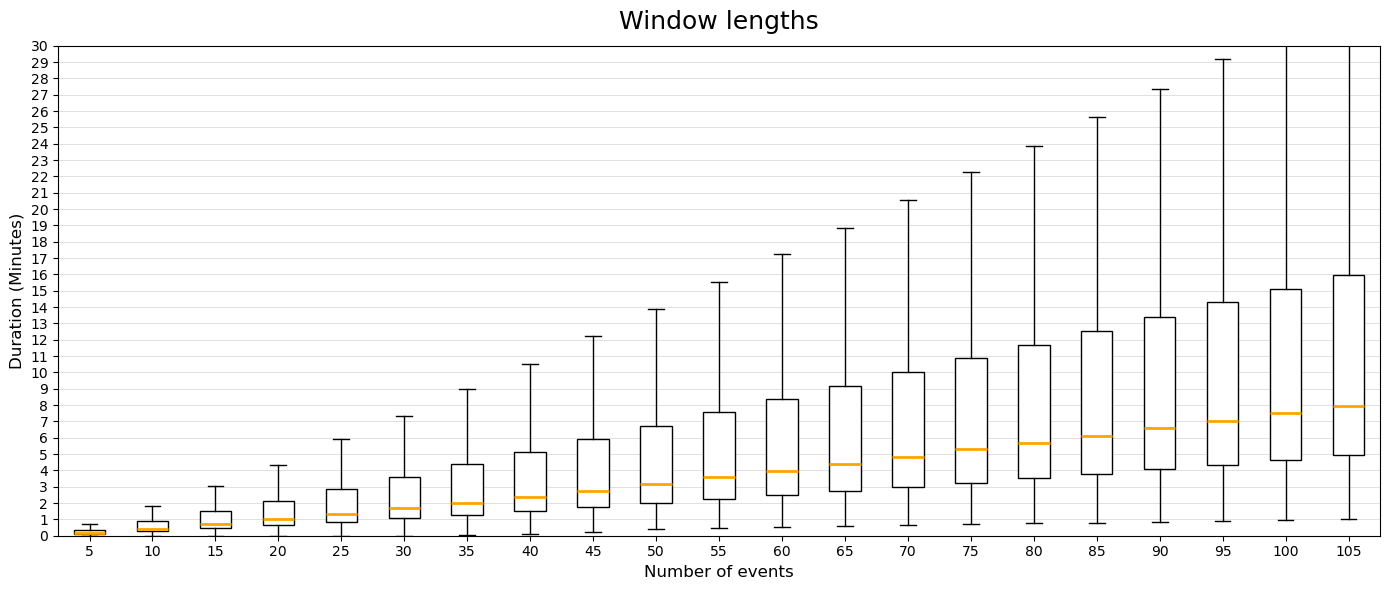}
    }
    \caption{Distribution of event-based windows timespan by varying the number of events $k$.}
    \label{fig:durations_comparison}
\end{figure*}

Following prior work in the literature \cite{krishnan2014activity}, we first considered a window length of $k=30$ events. 
Figures \ref{fig:aruba_durations} and \ref{fig:milan_durations} show the distribution of window timespan when changing the number of events ($k$). Note that the value $k=30$ corresponds to an average temporal span between 1 and 4-5 minutes; these are values coherent with those picked in time-based approaches for the same datasets~\cite{liciotti2020sequential,thukral2025layout}. 
Tests with lower and higher values of $k$ are reported in the Appendix, but we did not observe significant improvements.

The step parameter, instead, was fixed to $s = 10$ events. This choice was primarily motivated by computational cost considerations, as evaluating all maximally overlapping windows over long-duration sensor streams would result in an excessive number of LLM queries. The selected value represents a practical trade-off between temporal resolution and computational efficiency. Moreover, 
across both CASAS datasets used in our experiments, an average of 10 sensor events occurs within approximately 30 seconds. When modeling ADLs, this timespan can be reasonably assumed to correspond to a single ongoing activity. Consequently, advancing the window by $10$ events implies that the predicted activity label is held constant over this short interval, which doesn't significantly affect the evaluation. Moreover, in the rare cases where an activity transition occurs within this interval, the resulting mismatch leads to a pessimistic bias (i.e., delayed detection of activity changes), thereby yielding a conservative and fair assessment of model performance.

\subsection{Baselines}
We aim to assess the performance of our proposed system in comparison to the following baselines:

\begin{itemize}
    \item \textbf{\AdlLlmBase{}}: An LLM-based zero-shot ADL recognition method using time-based segmentation.
    \item \textbf{\AlSH{}}: A state-of-the-art machine learning method  using event-based segmentation. It transforms timestamped sensor data into event-based sliding windows and extracts handcrafted contextual and temporal features from each window. A Random Forest classifier is then trained to predict activity labels.
\end{itemize}

\subsection{ADL Recognition recognition rate}
\subsubsection{Impact of the LLM}
In this experiment, we evaluated the impact of different underlying LLMs on our event-based zero-shot approach. 
For this experiment, the temperature of the LLM has been set to 0 to reduce the variability between answers. This comparison highlights how model size and architecture affect recognition accuracy, as larger models generally capture richer contextual information, while smaller models offer advantages in terms of efficiency and deployment feasibility. Table \ref{tab:macro_weighted_llm_comparison} presents a comparison of the F1-scores achieved by our method across three different large language models. 

\begin{table}[t]
\centering
\caption{Weighted F1-score of the supervised \AlSH{} (upper bound) compared to our zero-shot approach.}
\label{tab:macro_weighted_vs_al_sh}
\begin{tabular}{clcc}
\toprule
& \textbf{\AlSH{}} & \textbf{Ours} \\
\midrule
\textbf{CASAS Aruba}
& 0.85 & 0.80 \\
\midrule
\textbf{CASAS Milan}
& 0.67 & 0.56\\
\bottomrule
\end{tabular}
\end{table}

As expected, the Gemma3 27B model outperforms the smaller Gemma3 4B variant, reflecting the advantages of increased model capacity. Notably, Gemma3 27B also surpasses the proprietary OpenAI model GPT-4o mini, underscoring the feasibility and competitiveness of locally deployed open-source models. The table reports F1-scores for each model on both evaluated datasets. More detailed results, including per-activity F1-scores and confusion matrices, are provided in the Appendix. In the following, all the experiments with our proposed method \acronym{} have been carried out with the best-performing model Gemma3 27B.

\begin{table}[t]
\centering
\caption{Weighted average F1-score of \acronym{} on the two datasets.}
\label{tab:macro_weighted_llm_comparison}
\begin{tabular}{lccc}
\toprule
 & \textbf{Gpt-4o-mini} & \textbf{Gemma3:4b} & \textbf{Gemma3:27b} \\
\midrule
\textbf{CASAS Aruba} & 0.78 & 0.76 & \textbf{0.80} \\
\textbf{CASAS Milan} & 0.51 & 0.52 & \textbf{0.56} \\
\bottomrule
\end{tabular}
\end{table}

\subsubsection{Comparison with time-based zero-shot (\AdlLlmBase{})}
In this experiment, we compare our event-based context approach \acronym{} with its a state-of-the-art approaches leveraring time-based segmentation~\AdlLlmBase{}. The results, shown in Table \ref{tab:macro_weighted_vs_adl-llm}, demonstrate that \acronym{} outperforms the baseline on both datasets. More detailed results, including per-activity F1-scores, are provided in the Appendix.

\begin{table}[t]
\centering
\caption{Comparison between the weighted average F1-score of \AdlLlmBase{} and our approach.}
\label{tab:macro_weighted_vs_adl-llm}
\begin{tabular}{clcccc}
\toprule
& \textbf{ADL-LLM} & \textbf{Ours} \\
\midrule
\textbf{CASAS Aruba}
& 0.50 & \textbf{0.80} \\
\midrule
\textbf{CASAS Milan}
& 0.40 & \textbf{0.56} \\
\bottomrule
\end{tabular}
\end{table}

\subsubsection{Comparison with supervised event-based approach (\AlSH{})}
In this experiment, we compare our zero-shot event-based approach \acronym{} with its supervised counterpart. The results are reported in Table \ref{tab:macro_weighted_vs_al_sh}. As expected, the supervised classifier outperforms the zero-shot model. As shown by the detailed results in the Appendix, this performance gap is particularly pronounced for activities that rely heavily on sequential patterns rather than on the semantic information of the involved sensors, such as Eating, Entering Home, or OTHER. The latter class is inherently not associated with any specific sensor and can therefore only be reliably identified through learned temporal dependencies. For this reason, the performance of the supervised model can be interpreted as an empirical upper bound. However, this bound is likely optimistic, as the supervised model likely overfits to household-specific temporal patterns and sensor configurations, and training on data from a given home and evaluating on a different one would lead to a substantial performance degradation. On the other hand, our zero-shot approach, while inferior in absolute terms, remains sufficiently competitive and potentially more robust to domain shifts.

\subsection{Confidence Impact}

In this experiment, we assessed the potential advantages of leveraging prediction repetition to derive a confidence measure. For this analysis, we set the number of repetitions $N = 5$. Moreover, we set the temperature of the LLM to 1 in order to increase the variability of the answers and maximally exploit the potential of the repetitions. We first compared the confidence scores of correct predictions against those of incorrect predictions. As can be seen from table \ref{tab:confidence_difference}, across all datasets and models, a consistent pattern emerged: correct predictions were associated with significantly higher confidence values, whereas incorrect predictions tended to exhibit lower confidence. This indicates that model confidence can serve as a reliable indicator of prediction quality in our setting.

\begin{table}[]
\centering 
\caption{Our model's confidence for correct predictions with respect to wrong predictions.}
\label{tab:confidence_difference}
\begin{tabular}{l cc cc cc }
\toprule
  & \multicolumn{2}{c}{\textbf{Gemma3:4b}} & \multicolumn{2}{c}{\textbf{Gemma3:27b}} \\ 
  \midrule
  & Correct & Wrong & Correct & Wrong \\ 
\midrule
\textbf{CASAS Aruba} & 0.92$\pm$0.02 & 0.78$\pm$0.04 & 0.94$\pm$0.02 & 0.87$\pm$0.03 \\
\textbf{CASAS Milan} & 0.94$\pm$0.02 & 0.84$\pm$0.04 & 0.97$\pm$0.01 &  0.90$\pm$0.03 \\ 
\midrule
\end{tabular}
\end{table}

To further investigate, we evaluated the recognition rate when we considered only the predictions whose confidence exceeded or met a specified threshold $th$, to evaluate (1) whether and how performance improves when only high-confidence predictions are used, and (2) the proportion of instances that would be discarded under such a filtering strategy. We tested four values for $th$: $0$ (i.e., confidence is not used), $0.66$, $0.8$, and $1$. The aggregated results are presented in Table \ref{tab:macro_weighted_thresholds_comparison}, with detailed results provided in the Appendix.

\begin{table}[h!]
\centering
\caption{Impact of confidence threshold $th$ on weighted average F1 score and the percentage of discarded instances ($N=5$).}
\label{tab:macro_weighted_thresholds_comparison}
\begin{tabular}{clcccc}
\toprule
 &  & \textit{th}=$0.00$ & \textit{th}=$0.66$ & \textit{th}=$0.80$ & \textit{th}=$1.0$ \\
\midrule
\multirow{2}{*}{\textbf{Gemma3:4b}} 
 & \textbf{F1-score} & 0.77 & 0.81 & 0.85 & \textbf{0.90} \\
 & \textbf{Discarded } & 0\% & 5\%  & 14\%  & 27\% \\
\midrule
\multirow{2}{*}{\textbf{Gemma3:27b}} 
 & \textbf{F1-score} & 0.79 & 0.80 & 0.86 & \textbf{0.90} \\
& \textbf{Discarded  } & 0\%  & 1\%  & 7\%  & 12\% \\
\bottomrule
\end{tabular}
\end{table}

Overall, the results are promising. Using confidence consistently improves performance. Remarkably, a smaller and lightweight model such as Gemma3 4B can reach performance levels comparable to Gemma3 27B, while the proportion of discarded instances remains sufficiently low so as not to undermine the overall effectiveness of the method.

%% file: sections/discussion.tex
\subsection{Hallucinations in LLMs}

Current LLMs are known to hallucinate. In our method, this can happen in three distinct areas: the classified activity, the reasoning process underlying the classification, and the output format. Format hallucinations, such as deviations from the expected JSON structure, occur frequently in small models like \textit{Gemma3:4B}, but are generally easy to detect and correct. Larger models rarely make format errors but may still hallucinate in classification and reasoning.

Classification hallucinations are the most directly impactful, as they affect the recognition rate (e.g., the F1 score). In practice, these are equivalent to mispredictions of traditional machine learning systems. 
The experiments reported in Table~\ref{tab:macro_weighted_thresholds_comparison} suggest that the confidence mechanism of \acronym{} has the potential to mitigate hallucinations, since the predictions that are discarded due to low confidence are more likely to be associated with hallucinations.

Reasoning hallucinations, on the other hand, carry a more subtle risk. While they may not always influence the correctness of classification, they can affect how users interpret the model's output. This may be problematic in cases where the classification is incorrect, but the reasoning appears plausible, leading users to mistakenly trust the system~\cite{fiori2025leveraging}. Such cases increase the risk of over-reliance and misplaced confidence, especially 
in contexts requiring interpretability or justification.

\subsection{Real-life deployment}
Our experimental evaluation considered several language models, including a cloud-based solution and locally hosted open-weights models. While cloud-based solutions do not require significant deployment efforts, they are costly, as they involve paying for each processed token. Moreover, their usage relies on the availability of a stable internet connection, introducing potential bottlenecks in terms of latency and throughput. Even more importantly, privacy concerns arise, since data transmitted to an untrusted third-party (e.g., OpenAI) may reveal personal activity habits.

The local deployment of open-weight models eliminates the aforementioned limitations, offering full control over data and ensuring privacy. However, this approach requires hardware with sufficient resources to run LLMs efficiently. In this work, we used Ollama to perform inference locally and tested two Gemma3 variants to explore the trade-off between performance and hardware requirements. 
%

Based on our experiments, the latency observed with the considered models is compatible with successfully applying event-based segmentation even for real-time ADL recognition.
In particular, the Gemma3:4b model required, on average, 3.1 ± 6.9 seconds to generate a response on our event-based windows, while Gemma3:27b model exhibited a slightly higher average response time of 7.3 ± 4.3 seconds (attributable to longer chain-of-thoughts leading to better results and a lower token-per-second throughput). 
While we also experimented with other open-weight reasoning models (e.g., DeepSeek-R1), we observed a higher latency for the reasoning process, and the generation 
of a longer explanation 
before providing the predicted activity. 
While these explanations may be valuable for explainability, minimising latency was more important.
Nonetheless, if only offline processing is required, the data may be processed at night when the volume of new events is minimal, leading to a more efficient use of resources. 



\subsection{Research Opportunities with LLM Confidence Estimation}
The introduction of a confidence estimation method for ADL classification using LLMs opens several interesting research directions. 

One possibility is to refine the model by using low-confidence predictions. For instance, when the LLM confidence is below a threshold, the system may trigger an active learning query to obtain from the user the ground truth about the performed activity. The collected labelled data could be used to improve ADL predictions with few-shot prompting, or they can be used to fine-tune the LLM and personalise it for a specific subject/environment.

Notably, the confidence value can also serve as a means for hallucination detection by identifying cases where the LLM is unsure about the predicted activity.
Hence, explainable AI (XAI) techniques could be enhanced by providing a quantifiable measure of certainty alongside explanations, thus making the LLM more trustworthy for end-users or domain experts.

%% file: sections/conclusion.tex
In this paper, we introduced a novel zero-shot method for ADL recognition with LLMs, leveraging event-based segmentation and confidence estimation. Our experiments on well-known public datasets show that the new approach significantly improves the state-of-the-art.

While our results are promising, this, as well as approaches in related work, would benefit from validation on data from more recent real-world deployments in home environments equipped with modern IoT devices. While such annotated public data is currently missing, in the context of a project in collaboration with neurologists, we are acquiring data from modern smart homes of subjects diagnosed with Mild Cognitive Impairment (MCI). In future work, we will evaluate the effectiveness of our methods on this data to support clinicians in the early diagnosis and progression of cognitive decline.

%% file: sections/appendix.tex

\appendix

\onecolumn

In the following, we present more detailed experimental results.
\begin{figure}[h]
    \centering
    \includegraphics[width=0.8\linewidth]{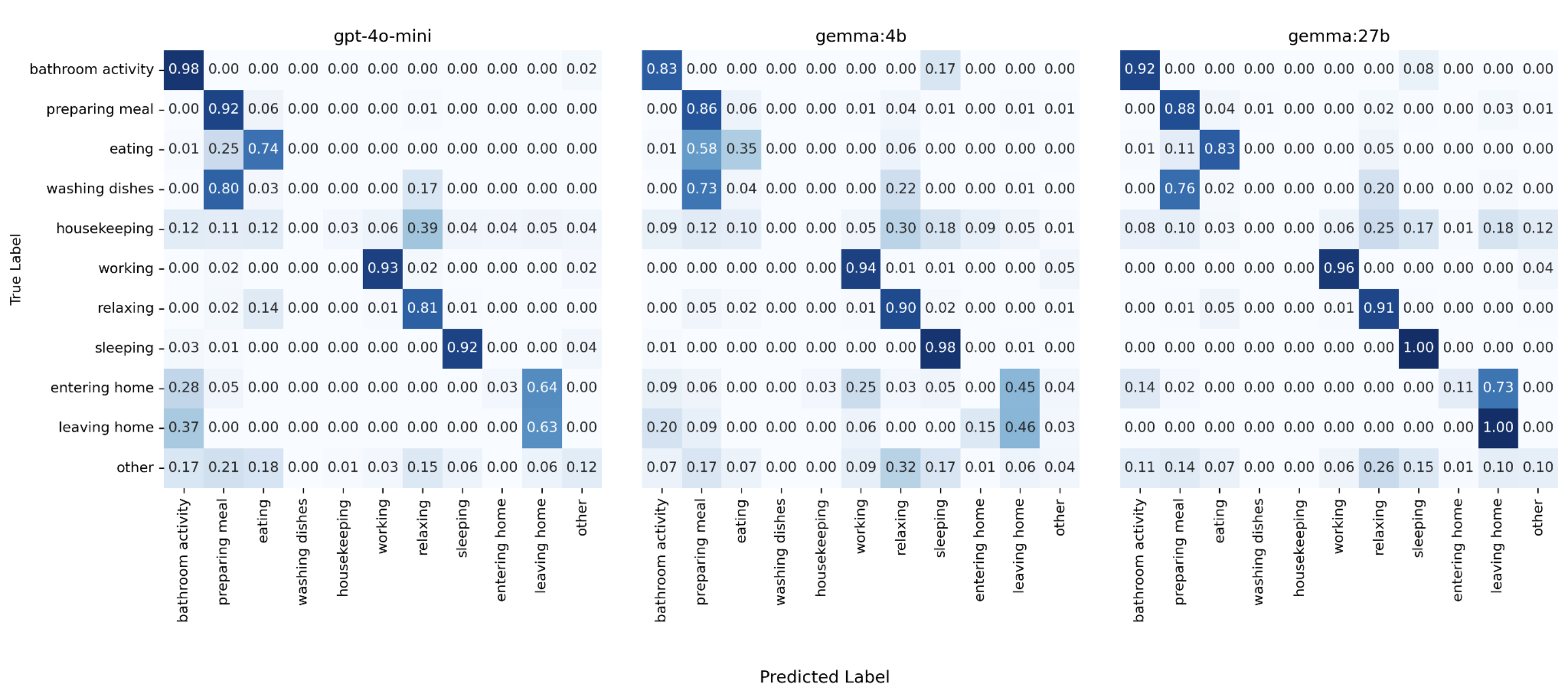}
    \caption{Confusion matrix for CASAS Aruba}
    \label{fig:cf_aruba}

    \includegraphics[width=0.8\linewidth]{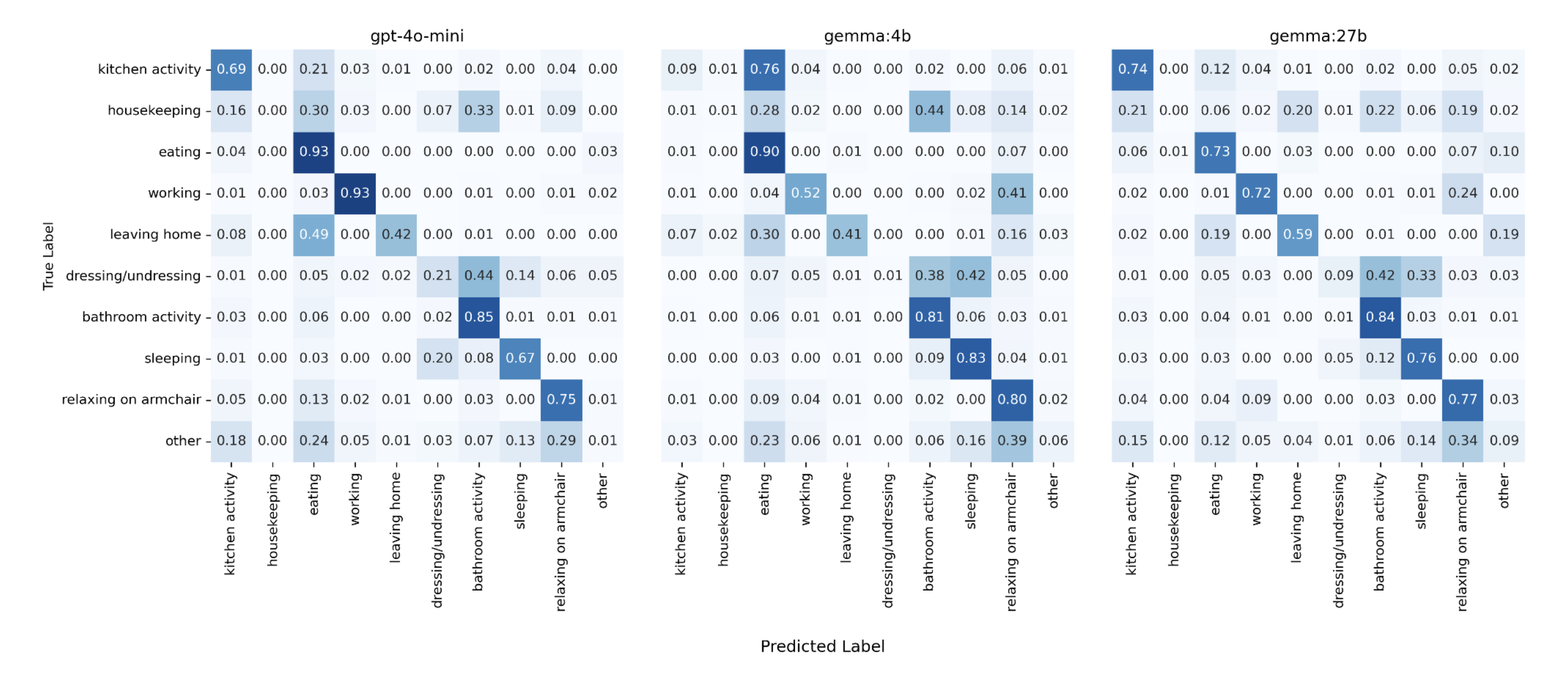}
    \caption{Confusion matrix for CASAS Milan}
    \label{fig:cf_milan}
\end{figure}

\vspace{1mm}

\begin{table}[h]
\centering
\tiny
\caption{F1-score for each activity class: CASAS Milan}
\label{tab:activity_results_milan}
\begin{tabular}{lccccc}
\toprule
\textbf{Activity} & \textbf{ADL-LLM} & \textbf{Gpt-4o-mini} & \textbf{Gemma3:4b} & \textbf{Gemma3:27b} & \textbf{AL} \\
\midrule
bathroom activity      & 0.47 & 0.48 & 0.45 & 0.44 & 0.51 \\
dressing/undressing    & 0.02 & 0.04 & 0.01 & 0.05 & 0.13 \\
eating                 & 0.05 & 0.05 & 0.04 & 0.07 & 0.00 \\
housekeeping           & 0.02 & 0.00 & 0.01 & 0.00 & 0.00 \\
kitchen activity       & 0.57 & 0.57 & 0.14 & 0.61 & 0.63 \\
leaving home           & 0.17 & 0.55 & 0.52 & 0.59 & 0.59 \\
other                  & 0.01 & 0.02 & 0.11 & 0.15 & 0.46 \\
relaxing on armchair   & 0.68 & 0.69 & 0.64 & 0.68 & 0.65 \\
sleeping               & 0.62 & 0.76 & 0.85 & 0.82 & 0.90 \\
working                & 0.55 & 0.50 & 0.23 & 0.26 & 0.50 \\
\midrule
Weighted avg           & 0.40 & 0.51 & 0.52 & 0.56 & 0.67 \\
\bottomrule
\end{tabular}
\end{table}

\begin{table}[h]
\centering
\tiny
\caption{F1-score for each activity class: CASAS Aruba}
\label{tab:activity_results_aruba}
\begin{tabular}{lccccc}
\toprule
\textbf{Activity} & \textbf{ADL-LLM} & \textbf{gpt-4o-mini} & \textbf{gemma3:4b} & \textbf{gemma3:27} & \textbf{AL} \\
\midrule
bathroom activity & 0.08 & 0.02 & 0.03 & 0.04 & 0.80 \\
eating            & 0.05 & 0.06 & 0.08 & 0.14 & 0.55 \\
entering home     & 0.00 & 0.01 & 0.00 & 0.02 & 0.82 \\
housekeeping      & 0.00 & 0.02 & 0.00 & 0.00 & 0.00 \\
leaving home      & 0.24 & 0.66 & 0.52 & 0.84 & 0.62 \\
other             & 0.03 & 0.18 & 0.07 & 0.18 & 0.77 \\
preparing meal    & 0.51 & 0.48 & 0.47 & 0.61 & 0.45 \\
relaxing          & 0.78 & 0.83 & 0.81 & 0.84 & 0.92 \\
sleeping          & 0.77 & 0.95 & 0.96 & 0.98 & 0.96 \\
washing dishes    & 0.00 & 0.00 & 0.00 & 0.00 & 0.00 \\
working           & 0.53 & 0.55 & 0.32 & 0.47 & 0.62 \\
\midrule
Weighted avg      & 0.50 & 0.78 & 0.76 & 0.80 & 0.85 \\
\bottomrule
\end{tabular}
\end{table}

\begin{table}[h]
\centering
\tiny
\caption{Performance comparison for different values of $k$ on the CASAS Aruba dataset (Gemma3-27b)}
\label{tab:k_comparison}
\begin{tabular}{lccccc}
\toprule
\textbf{Activity} & \textbf{$k=5$} & \textbf{$k=10$} & \textbf{$k=20$} & \textbf{$k=30$} & \textbf{$k=50$} \\
\midrule
bathroom activity & 0.03 & 0.06 & 0.03 & 0.04 & 0.02 \\
eating            & 0.10 & 0.14 & 0.11 & 0.14 & 0.13 \\
entering home     & 0.00 & 0.00 & 0.00 & 0.02 & 0.00 \\
housekeeping      & 0.00 & 0.00 & 0.01 & 0.00 & 0.00 \\
leaving home      & 0.52 & 0.71 & 0.68 & 0.84 & 0.64 \\
other             & 0.31 & 0.33 & 0.29 & 0.18 & 0.26 \\
preparing meal    & 0.46 & 0.64 & 0.64 & 0.61 & 0.55 \\
relaxing          & 0.77 & 0.87 & 0.87 & 0.84 & 0.85 \\
sleeping          & 0.87 & 0.98 & 0.98 & 0.98 & 0.96 \\
washing dishes    & 0.00 & 0.00 & 0.00 & 0.00 & 0.00 \\
working           & 0.41 & 0.57 & 0.54 & 0.47 & 0.56 \\
\midrule
Weighted avg      & 0.72 & 0.83 & 0.82 & 0.80 & 0.80 \\
\bottomrule
\end{tabular}
\end{table}

\begin{table}[h]
\centering
\tiny
\caption{Performance comparison for different threshold values on CASAS Aruba dataset (Gemma3-27b)}
\label{tab:threshold_comparison_v2}
\begin{tabular}{lcccc}
\toprule
\textbf{Activity} & \textit{th}=$0.00$ & \textit{th}=$0.66$  & \textit{th}=$0.80$  & \textit{th}=$1.00$  \\
\midrule
bathroom act   & 0,04 & 0,04 & 0,12 & 0,23 \\
eating          & 0,15 & 0,15 & 0,20 & 0,32 \\
entering home   & 0,00 & 0,00 & 0,00 & 0,00 \\
housekeeping    & 0,00 & 0,00 & 0,00 & 0,00 \\
leaving home    & 0,73 & 0,73 & 0,89 & 0,88 \\
other           & 0,12 & 0,12 & 0,19 & 0,33 \\
preparing meal  & 0,63 & 0,64 & 0,67 & 0,68 \\
relaxing        & 0,86 & 0,86 & 0,91 & 0,95 \\
sleeping        & 0,97 & 0,98 & 0,98 & 0,99 \\
washing dishes  & 0,00 & 0,00 & 0,00 & 0,00 \\
working         & 0,53 & 0,53 & 0,61 & 0,74 \\
\midrule
Weighted avg & 0,79 & 0,80 & 0,86 & 0,90 \\
\bottomrule
\end{tabular}
\end{table}

\begin{table}[h]
\centering
\tiny
\caption{Performance comparison for different threshold values on CASAS Aruba dataset (Gemma3-4b)}
\label{tab:threshold_comparison}
\begin{tabular}{lcccc}
\toprule
\textbf{Activity} & \textit{th}=$0.00$ & \textit{th}=$0.66$  & \textit{th}=$0.80$  & \textit{th}=$1.00$  \\
\midrule
bathroom act   & 0,08 & 0,08 & 0,13 & 0,21 \\
eating          & 0,12 & 0,13 & 0,19 & 0,19 \\
entering home   & 0,00 & 0,00 & 0,00 & 0,00 \\
housekeeping    & 0,00 & 0,00 & 0,00 & 0,00 \\
leaving home    & 0,59 & 0,60 & 0,68 & 0,68 \\
other           & 0,10 & 0,10 & 0,10 & 0,12 \\
preparing meal  & 0,51 & 0,54 & 0,61 & 0,71 \\
relaxing        & 0,82 & 0,84 & 0,89 & 0,92 \\
sleeping        & 0,97 & 0,98 & 0,98 & 0,98 \\
washing dishes  & 0,00 & 0,00 & 0,00 & 0,00 \\
working         & 0,39 & 0,61 & 0,68 & 0,69 \\
\midrule
Weighted avg    & 0,77 & 0,81 & 0,85 & 0,90 \\
\bottomrule
\end{tabular}
\end{table}